\pdfoutput=1

\documentclass[11pt]{article}
\usepackage[]{acl}

\def\vs{{\textbf{s}}}
\def\vt{{\textbf{t}}}

\def\vx{{\textbf{x}}}
\def\vy{{\textbf{y}}}

\usepackage{times}
\usepackage{latexsym}
\usepackage{array}
\usepackage{multirow}
\usepackage{multicol}
\usepackage{graphicx}
\usepackage{booktabs}
\usepackage{subcaption}
\usepackage{makecell}

\usepackage[T1]{fontenc}

\usepackage[utf8]{inputenc}

\usepackage{microtype}
\title{Leveraging Synthetic Targets for Machine Translation}

\author{Sarthak Mittal\Thanks{$\;$Work done during an internship at NVIDIA. Correspondence author \href{mailto:sarthmit@gmail.com}{sarthmit@gmail.com}}$^{\;\;1,2}$ \And Oleksii Hrinchuk$^3$ \And Oleksii Kuchaiev$^3$ \\[5pt]
\hspace{-105mm}$^1$Mila, $^2$Universite de Montreal, $^3$NVIDIA
}
\begin{document}
\maketitle
\begin{abstract}
In this work, we provide a recipe for training machine translation models in a limited resource setting by leveraging synthetic target data generated using a large pre-trained model. We show that consistently across different benchmarks in bilingual, multilingual, and speech translation setups, training models on synthetic targets outperforms training on the actual ground-truth data. This performance gap grows bigger with increasing limits on the amount of available resources in the form of the size of the dataset and the number of parameters in the model. We also provide preliminary analysis into whether this boost in performance is linked to ease of optimization or more deterministic nature of the predictions, and whether this paradigm leads to better out-of-distribution performance across different testing domains.
\end{abstract}

\vspace{-1mm}
\section{Introduction}
\vspace{-1mm}
Neural Machine Translation (NMT)\let\thefootnote\relax\footnotetext{We used the NeMo codebase \cite{kuchaiev2019nemo} for all our experiments.} \cite{bahdanau2014neural,wu2016google,stahlberg2020neural} relies on deep learning models to train end-to-end translation systems. With the advent of deep recurrent models like LSTMs~\cite{hochreiter1997long,sundermeyer2014translation,chung2014empirical} and their attention-augmented improvements~\cite{bahdanau2014neural,luong2015effective}, these models outperformed traditional statistical~\cite{koehn2009statistical,della1994mathematics} and rule-based~\cite{lagarda2009statistical,nirenburg1989knowledge} approaches. Recently, with the introduction of fully attention-based networks networks~\cite{vaswani2017attention,dehghani2018universal,sukhbaatar2019adaptive,dai2019transformer,kitaev2020reformer,choromanski2020rethinking,mittal2021compositional} and increase in compute and data, large-scale Transformer networks have dominated the field of natural language processing~\cite{devlin2018bert,brown2020language,hoffmann2022training,shoeybi2019megatron}, and machine translation in particular~\cite{edunov2018understanding,raffel2020exploring}, leading to not only better performance but also more efficient training through their parallel computations~\cite{ott-etal-2018-scaling}.

While it has been established that scaling up data and compute boosts the performance of the large-scale NMT systems \cite{gordon2021data,ghorbani2021scaling,kaplan2020scaling,bahri2021explaining}, there is still a need to focus on budget models that can run on mobile and edge computing devices. In other tasks, like end-to-end speech translation, training data is scarce and expensive. Inspired from these needs, we provide a recipe for training text-to-text and text-to-speech translation systems in a limited resource setting at the modest overhead of running inference of pre-trained models on the source sentences.

Though in theory, increasing the amount of data provides a relatively simple methodology for bolstering the performance of current AI systems, it is difficult to do so when obtaining new data is costly, especially because of the labeling process in supervised learning. 
On the other hand, there have been a variety of approaches leveraging synthetic data to either improve the robustness of the systems or to boost their performance. This can be achieved by introducing adversarial examples in the training set \cite{goodfellow2014explaining}, considering knowledge distillation when provided access to large models but not their pre-training data~\cite{bucilua2006model,gou2021knowledge,hinton2015distilling,kim2016sequence,urner2011access,cheng2020explaining,phuong2019towards,tan2019multilingual}, or using forward and back translation techniques \cite{zhang2016exploiting,sennrich2015improving,bogoychev2019domain,edunov2018understanding,hoang2018iterative} when additional monolingual data, which is easily available, is leveraged to generate synthetic targets to augment the amount of data.

In this work, we use large, often pre-trained, NMT systems to provide synthetic targets which can be leveraged to train high performing low-compute and low-data models. Our findings show that using these synthetic translations to train different NMT systems leads to considerable improvements, even if we remove all ground-truth feedback from the training set. We also test models trained with synthetic targets on out-of-distribution settings through translations on different domains in a zero-shot manner as well as finetuning them on a different domain using additional synthetic data from an existing finetuned model, and further highlight the improvements obtained in both. We also find additional evidence to support that the ease of optimization resulting from training on synthetic targets does not completely explain its superior performance \cite{he2019revisiting}. Instead, we showcase that models trained with synthetic targets are more deterministic than their counterparts which are trained on real targets.

Our key contributions are as follows
\begin{itemize}
    \item We provide a recipe for training better models in resource and compute constrained settings provided access to a pre-trained system and validate it on bilingual and multilingual translation setups, as well as out-of-domain generalization.
    \item We provide analysis into the reasoning behind this improved performance and demonstrate that it is not solely because of ease of optimization but instead, we believe it is due to more deterministic nature of such systems.
\end{itemize}
\vspace{-1mm}
\section{Related Work}
\vspace{-1mm}
Synthetic targets have been consistently used in the past to either augment the amount of data available or to boost performance through knowledge transfer between models. In the domain of Machine Translation, a popular way of augmenting data is by considering monolingual data that is available in abundance and obtaining its paired counterpart using a trained model. 

\textbf{Back Translation}. Back translation \cite{sennrich2015improving,edunov2018understanding,hoang2018iterative} relies on translating unseen sentences from the target side back to the source side using an existing trained model to provide new paired samples which can then be leveraged to train a model in the forward direction.. For example, if the task is to translate sentences from language $\mathcal{S}$ to $\mathcal{T}$ ($\mathcal{S}\to\mathcal{T}$), one can obtain a corpus of monolingual data from $\mathcal{T}$ and translate it backwards to $\mathcal{S}$ using an existing trained $\mathcal{T}\to\mathcal{S}$ translation model. This would then provide additional paired data that can be combined with the existing ground-truth data to train a $\mathcal{S}\to\mathcal{T}$ translation model.

\textbf{Forward Translation}. Analogous to back translation, forward translation \cite{zhang2016exploiting,bogoychev2019domain} or self-training~\cite{he2019revisiting} relies on training a standard translation model using which additional data is obtained by translating new sentences from the source to the target side and then re-training the translation model using this additional data. For example, to translate in the $\mathcal{S}\to\mathcal{T}$ direction, one first trains a model using existing data and then leverages this model to generate targets for a corpus of monolingual data from $\mathcal{S}$ to provide new paired data. This data is then combined with the original data for re-training of the $\mathcal{S}\to\mathcal{T}$ translation model. Typically, forward translation is not as effective as back translation since the errors of the model are further propagated in the data in the former case \cite{burlot2019using}.

Our approach can also be related to knowledge distillation, which has been the popular choice for transferring knowledge from a larger (called teacher) to a smaller model (called student) by enforcing similarities at different levels \cite{hinton2015distilling,gou2021knowledge,freitag2017ensemble,kim2016sequence}, eg. in the output or the representation space of the two models. We give a brief overview of the different strategies used.

\textbf{Soft Target Matching}. The earliest works on knowledge distillation transfer knowledge by enforcing the soft-logits of the student to be close to those of the teacher \cite{hinton2015distilling}. This is accomplished by introducing a loss term that penalizes deviation of the student model's logits from the teacher, and this loss can either be used as is or added as a regularizing effect in training. This formulation of knowledge distillation has been revisited as \textit{Word-Level Knowledge Distillation} for NMT in \cite{kim2016sequence,wang2021selective}. While it is a simple way of distilling the teacher's knowledge into the student, it can be computationally expensive if the number of classes, or equivalently the vocabulary $\mathcal{V}$, is large as it requires either storing all the $|\mathcal{V}|$ soft-logits for all words of the whole dataset or requires access to the teacher model on the fly, which can make training slow.

\begin{table}[t!]
    \centering
    \small
    \def\arraystretch{1.25}
    \begin{tabular}{ccccc}
    \toprule
    \textbf{Model} & \textbf{Data} & \textbf{Real} & \textbf{Synthetic} & $\mathbf{\Delta}$ \\
    \midrule
    \multirow{3}{*}{$2\times 2$} & $0.5$M & $17.6$ & $20.9$ & $3.3$ \\
     & $5$M & $22.6$ & $23.5$ & $0.9$ \\
     & $25$M & $23.1$ & $23.9$ & $0.8$ \\
     \midrule
    \multirow{3}{*}{$6\times 6$} & $0.5$M & $18.0$ & $22.5$ & $4.5$ \\
     & $5$M & $24.7$ & $25.7$ & $1.0$ \\
     & $25$M & $25.3$ & $25.9$ & $0.6$ \\
     \midrule
    \multirow{3}{*}{$24\times 6$} & $0.5$M & $18.6$ & $22.9$ & $4.3$ \\
     & $5$M & $25.7$ & $26.5$ & $0.8$ \\
     & $25$M & $26.4$ & $26.4$ & $0.0$ \\
     \bottomrule
    \end{tabular}
    \caption{English to Russian NMT models trained on subsets of WMT21 Machine Translation dataset. Synthetic targets are generated with $24\times6$ teacher model trained on the full $25$M WMT dataset.}
    \label{tab:en-ru}
    \vspace{-5mm}
\end{table}

\textbf{Representation Matching}.\cite{romero2014fitnets,zagoruyko2016paying,lee2018self,heo2019knowledge,passban2021alp,chen2021cross} Another way of transferring knowledge from the teacher to the student is by matching their intermediate representations. Again, this can be accomplished by considering a regularization term $l_{reg}(g_t(\phi_t), g_s(\phi_s))$ where $l_{reg}$ is some notion of similarity, $\phi_t,\phi_s$ are the intermediate representations of the teacher and student respectively and $g_t,g_s$ are functions that map the two representations to the same space, which is needed as the student is often smaller than the teacher model. While intuitively simple, this formulation is harder to implement and tune as the intermediate representations may be of different shapes, making it non-trivial to obtain a notion of similarity between the two. For example, if $g_t$ and $g_s$ map all the representations to the same point, the matching loss would be low even though the representations themselves $\phi_t, \phi_s$ can be quite dis-similar.

\textbf{Sequence-Level Knowledge Distillation}. \citet{kim2016sequence} propose \textit{Sequence-Level Knowledge Distillation} which does not rely on soft-logits from the teacher model but instead relies on the synthetic translations obtained from the teacher model.
Using synthetic targets is computationally efficient as the computation does not rely on matching the soft-logits across the whole of vocabulary but instead relies on sparse signals. Moreover, \citet{kim2016sequence} showcase that using synthetic targets in LSTM-based systems lead to improved performance as opposed to the traditional knowledge distillation approach based on matching soft-logits.

\begin{table}[t!]
    \centering
    \small
    \def\arraystretch{1.25}
    \begin{tabular}{ccccc}
    \toprule
    \textbf{Model} & \textbf{Data} & \textbf{Real} & \textbf{Synthetic} & $\mathbf{\Delta}$ \\
    \midrule
    \multirow{3}{*}{$2\times 2$} & $0.5$M & $18.6$ & $22.8$ & $4.2$ \\
     & $5$M & $23.8$ & $26.0$ & $2.2$ \\
     & $57$M & $24.2$ & $26.3$ & $2.1$ \\
     \midrule
    \multirow{3}{*}{$6\times 6$} & $0.5$M & $18.7$ & $23.9$ & $5.2$ \\
     & $5$M & $25.1$ & $27.8$ & $2.7$ \\
     & $57$M & $26.6$ & $28.4$ & $1.8$ \\
     \midrule
    \multirow{3}{*}{$24\times 6$} & $0.5$M & $19.2$ & $24.1$ & $4.9$ \\
     & $5$M & $25.9$ & $28.4$ & $2.5$ \\
     & $57$M & $27.1$ & $29.0$ & $1.9$ \\
     \bottomrule
    \end{tabular}
    \caption{English to German NMT models trained on subsets of WMT21 Machine Translation dataset. Synthetic targets are generated with $24\times6$ teacher model trained on large amounts of data in addition to the full $57$M WMT training dataset.}
    \label{tab:en-de}
    \vspace{-6mm}
\end{table}

\begin{table*}[t!]
    \centering
    \small
    \def\arraystretch{1.25}
    \begin{tabular}{c| c | c c c c | c c c c | c}
    \toprule
    \multirow{2}{*}{\textbf{Model}} & \multirow{2}{*}{\textbf{Data}} & \multicolumn{4}{c|}{\textbf{Real}} & \multicolumn{4}{c|}{\textbf{Synthetic}} & $\mathbf{\Delta}$ \\
    & & \textbf{en-de} & \textbf{en-es} & \textbf{en-fr} & \textbf{Avg.} & \textbf{en-de} & \textbf{en-es} & \textbf{en-fr} & \textbf{Avg.} & \textbf{Avg.} \\
    \midrule
    \multirow{3}{*}{$2\times2$} & $1.5$M & $24.9$ & $29.6$ & $29.3$ & $27.9$ & $28.2$ & $31.4$ & $31.3$ & $30.3$ & $2.4$ \\
    & $15$M & $29.5$ & $31.8$ & $32.6$ & $31.3$ & $31.9$ & $33.0$ & $33.5$ & $32.8$ & $1.5$ \\
    & $300$M & $30.1$ & $32.5$ & $32.8$ & $31.8$ & $32.3$ & $33.2$ & $33.7$ & $33.0$ & $1.2$ \\
     \midrule
    \multirow{3}{*}{$6\times6$} & $1.5$M & $25.9$ & $30.1$ & $30.2$ & $28.7$ & $31.0$ & $32.5$ & $32.5$ & $32.0$ & $3.3$ \\
    & $15$M & $33.2$ & $33.9$ & $35.0$ & $34.0$ & $36.0$ & $34.4$ & $35.4$ & $35.2$ & $1.2$ \\
     & $300$M & $34.5$ & $34.4$ & $35.6$ & $34.8$ & $36.1$ & $34.5$ & $35.8$ & $35.4$ & $0.6$ \\
     \midrule
    \multirow{3}{*}{$24\times6$} & $1.5$M & $25.8$ & $30.0$ & $29.6$ & $28.5$ & $31.8$ & $33.3$ & $33.3$ & $32.8$ & $4.3$ \\
    & $15$M & $35.0$ & $34.5$ & $35.7$ & $35.0$ & $36.9$ & $34.9$ & $36.4$ & $36.0$ & $1.0$ \\
     & $300$M & $36.1$ & $34.9$ & $36.8$ & $35.9$ & $37.9$ & $35.2$ & $37.0$ & $36.7$ & $0.8$ \\
     \bottomrule
    \end{tabular}
    \caption{Multilingual Machine Translation systems trained to translate sentences from English to three different languages: German, Spanish and French. Synthetic targets are generated with $24\times6$ bilingual teachers trained on a large corpus of parallel data. SacreBLEU on WMT20 dev set is reported.}
    \label{tab:en-mult}
    \vspace{-5mm}
\end{table*}

While similar to forward translation and sequence-level knowledge distillation, our approach differs by leveraging pre-trained translation models trained on large amounts of data for synthetic targets as opposed to training from scratch and then re-training. Further, we also consider setups where the amount of data used for the teacher and the student model is different, and where their model sizes can be similar.
\vspace{-1mm}
\section{Method}
\vspace{-1mm}
Our aim is to perform Machine Translation from a source language $\mathcal{S}$ to a target language $\mathcal{T}$ given some training data $\mathcal{D}_{\mathcal{S}\to\mathcal{T}}=\{(\vs_i, \vt_i)\}_{i=1}^N$ where $\vs_i \in \mathcal{S}$ is the source sentence and $\vt_i \in \mathcal{T}$ denotes the target sentence which is the ground-truth translation corresponding to $\vs_i$. Further, we assume that we have access to a teacher network $f_{\mathcal{S}\to\mathcal{T}}(\cdot)$, using which we obtain synthetic targets $f_{\mathcal{S}\to\mathcal{T}}(\vs_i)$ to construct the synthetic dataset $\mathcal{D}'_{\mathcal{S}\to\mathcal{T}}=\{(\vs_i, f_{\mathcal{S}\to\mathcal{T}}(\vs_i))\}_{i=1}^N$.

For our experiments, we consider different dataset sizes for the student models by subsampling from $\mathcal{D}_{\mathcal{S}\to\mathcal{T}}$ and $\mathcal{D}'_{\mathcal{S}\to\mathcal{T}}$ respectively. All the models considered in this work rely on the Encoder-Decoder Transformer architecture \cite{vaswani2017attention} with the teacher network generally having 24 encoder and 6 decoder layers ($24\times 6$). We consider different model sizes for the student, ranging from small models to matching the teacher's size.

Typically, knowledge distillation considers the same input data for training both the teacher and the student. Instead, we perform analysis where the student has access to different amounts of data. Also, unlike knowledge distillation where knowledge is transferred from a bigger teacher to a smaller student network, we additionally consider setups where the student network is as big as the teacher, and showcase the benefits in this regime.
\vspace{-1mm}
\section{Experiments}
\vspace{-1mm}
\label{sec:exp}
For all our experiments, we only considered Transformer models for both teachers and students. Unless specified, we used the Pre-Layer Normalization variant where LayerNorm is applied before the respective attention and residual computations~\cite{xiong2020layer}. In our experiments, we consider two text-to-text machine translation setups: bilingual and multilingual, and one speech-to-text setup. For the text-to-text machine translation experiments, we considered byte-pair encoding \cite{britz2017massive} for bilingual experiments and sentence-piece encoding \cite{kudo2018sentencepiece} for multilingual experiments.

\begin{table*}[t]
    \centering
    \small
    \def\arraystretch{1.25}
    \begin{tabular}{c| c | c c c c | c c c c | c}
    \toprule
    \multirow{2}{*}{\textbf{Model}} & \multirow{2}{*}{\textbf{Data}} & \multicolumn{4}{c|}{\textbf{Real}} & \multicolumn{4}{c|}{\textbf{Synthetic}} & $\mathbf{\Delta}$ \\
    & & \textbf{de-en} & \textbf{es-en} & \textbf{fr-en} & \textbf{Avg.} & \textbf{de-en} & \textbf{es-en} & \textbf{fr-en} & \textbf{Avg.} & \textbf{Avg.} \\
    \midrule
    \multirow{3}{*}{$2\times2$} & $1.5$M & $28.4$ & $29.1$ & $32.2$ & $29.9$ & $31.9$ & $30.8$ & $34.1$ & $32.3$ & $2.4$ \\
    & $15$M & $34.6$ & $32.8$ & $36.2$ & $34.5$ & $37.1$ & $33.6$ & $37.5$ & $36.1$ & $1.6$ \\
    & $300$M & $35.5$ & $32.7$ & $36.9$ & $35.0$ & $38.3$ & $33.8$ & $37.6$ & $36.6$ & $1.6$ \\
     \midrule
    \multirow{3}{*}{$6\times6$} & $1.5$M & $30.8$ & $30.1$ & $33.2$ & $31.4$ & $35.3$ & $31.8$ & $35.3$ & $34.1$ & $2.7$ \\
    & $15$M & $38.1$ & $34.0$ & $38.3$ & $36.8$ & $40.2$ & $35.0$ & $39.1$ & $38.1$ & $1.3$ \\
    & $300$M & $40.0$ & $34.2$ & $38.8$ & $37.7$ & $41.3$ & $35.5$ & $39.6$ & $38.8$ & $1.1$ \\
     \midrule
    \multirow{3}{*}{$24\times6$} & $1.5$M & $31.4$ & $30.5$ & $33.8$ & $31.9$ & $36.0$ & $32.8$ & $36.2$ & $35.0$ & $3.1$ \\
    & $15$M & $38.8$ & $34.6$ & $39.6$ & $37.6$ & $40.8$ & $35.6$ & $40.1$ & $38.8$ & $1.2$ \\
    & $300$M & $40.1$ & $35.6$ & $40.9$ & $38.9$ & $41.8$ & $36.3$ & $41.2$ & $39.8$ & $0.9$ \\
     \bottomrule
    \end{tabular}
    \caption{Multilingual Machine Translation systems trained to translate sentences from German, Spanish, and French to English. Synthetic targets are generated with $24\times6$ bilingual teachers trained on a large corpus of parallel data. SacreBLEU on WMT20 dev set is reported.}
    \label{tab:mult-en}
    \vspace{-5mm}
\end{table*}

\vspace{-1mm}
\subsection{Bilingual Machine Translation}
We first test the benefits of training with synthetic targets on bilingual machine translation where models are trained to translate sentences from one specified (source) language to another (target) language. We conducted experiments with the source language as English and the target languages as Russian and German. We consistently see improvements when training with synthetic targets, and these benefits are substantial when the student is trained on limited data. Even on the same amount of data, we see benefits of using synthetic data when the student has lower complexity.

\textbf{English to Russian}. We used the WMT'21 dataset for English to Russian Machine Translation, where we trained the models with different tokenizers and vocabularies on the source and target side with the vocabulary size of 24576. For the teacher model, we trained a baseline Transformer with 24 encoder and 6 decoder layers. For the student, we consider two different axis of analysis: models with lower capacity than the $24 \times 6$ teacher, and models trained with fewer data subsampled from the training set used to train the teacher.

In Table \ref{tab:en-ru}, we highlight that in the low-data regime for the student, consistently across all the different model sizes, training solely on synthetic targets obtained from the teacher model leads to much better performance than training on real ground-truth data. We also see that even when keeping the amount of data fixed (25M sentences; which is the same data on which the teacher model was trained), we see improvements on using synthetic targets in training smaller models. Thus, the only avenue where we don't get a substantial improvement is when the student uses the same dataset as the teacher model and has high complexity, similar to the teacher. This, however, is intuitive and doesn't pose a problem since our aim is for better low-compute models.

\textbf{English to German}. We consider the WMT'21 dataset for the task of English to German Machine Translation, where we trained the models with shared tokenizers and vocabularies on the source and target side, with the vocabulary size of 32000. For the teacher model, we picked a published transformer model with 24 encoder and 6 decoder layers from the NeMo codebase \cite{kuchaiev2019nemo} which was trained on substantially more data than WMT. For the student models, we again consider two different axis of analysis: models with lower capacity than the $24 \times 6$ teacher, and models trained with different percentages of the WMT data.

In Table \ref{tab:en-de}, we see that consistent with our English to Russian experiments, in low data regime the student models trained with synthetic targets outperform the models trained on real ground-truth data. In particular, since the teacher model was trained on an enormous corpus outside of WMT as well, we see that even on using a $24 \times 6$ transformer model which has the same complexity as the teacher on the full WMT21 dataset (57M sentences), it is still beneficial to train it with synthetic data as opposed to real data. Additionally we can also see that smaller models ($6\times 6$) trained on synthetic targets outperform larger models ($24 \times 6$) trained on real targets, while also providing faster training and inference.

Our English to Russian and English to German experiments show that indeed across different amounts of data regime and model complexities, as long as one does not approach the large-scale teacher model in both the regimes, it is beneficial to train systems using synthetic targets from the pre-trained large-scale model as opposed to training it on the ground-truth targets. We show that through this simple recipe, one can train much better smaller scale models which is beneficial to have when considering deployment in low-resource settings where inference latency needs to be low. It also provides a viable strategy for training of models when access to large-scale systems is provided but their training data is not.

\begin{table*}[]
    \centering
    \small
    \def\arraystretch{1.25}
    \begin{tabular}{l|c|cc|ccccccc|c}
    \toprule
    \multirow{2}{*}{\textbf{German targets source}} & \multirow{2}{*}{\textbf{Size}} &\multicolumn{2}{c|}{\textbf{Must-C v2}} & \multicolumn{8}{c}{\textbf{IWSLT tst}} \\
    & & \textbf{dev} & \textbf{tst} & \textbf{2010} & \textbf{2013} & \textbf{2014} & \textbf{2015} & \textbf{2018} & \textbf{2019} & \textbf{2020} & \textbf{Avg.} \\
    \midrule
    Real & $590$K & $25.2$ & $27.8$ & $22.1$ & $27.9$ & $23.8$ & $21.3$ & $20.9$ & $20.1$ & $20.9$ & $22.4$ \\
    \midrule
    Synthetic, WMT21 teacher & $590$K & $28.3$ & $28.9$ & $24.5$ & $28.0$ & $23.9$ & $23.0$ & $22.6$ & $21.7$ & $23.3$ & $23.9$ \\
    \quad + fine-tuned on IWSLT & $590$K & $29.2$ & $30.2$ & $25.0$ & $29.5$ & $24.8$ & $24.9$ & $24.2$ & $23.4$ & $25.3$ & $25.3$ \\
    \quad\quad + extra ASR data & $1.25$M & $30.6$ & $31.0$ & $27.2$ & $31.3$ & $27.4$ & $25.8$ & $25.1$ & $24.3$ & $26.4$ & $26.8$ \\
    \bottomrule
    \end{tabular}
    \caption{Speech Translation systems trained to translate English audio to German text. Synthetic targets are generated with $24\times6$ bilingual teacher trained on WMT21 and fine-tuned on IWSLT22.}
    \label{tab:st}
    \vspace{-5mm}
\end{table*}

\vspace{-1mm}
\subsection{Multilingual Machine Translation}
Next, we move our attention towards multilingual machine translation where a single model is trained to perform translation from multiple different source languages to various different target languages. In particular, we focus on two different multilingual settings; translating sentences from (a) English to German, Spanish and French, and (b) German, Spanish and French to English. An important difference from bilingual experiments is that in this case, we obtain synthetic targets for the student multilingual models using published bilingual teachers. As in the bilingual setup, we again see consistent improvements of using synthetic targets instead of real ground-truth targets.

\textbf{English to German/Spanish/French}. We consider the setup of training a single model to translate english sentences into three different languages: german, spanish and french. These models are trained with shared tokenizers and vocabularies on the source and target side, with the vocabulary size of 64000. We query bilingual teachers trained on considerably large datasets to obtain synthetic targets for training the multilingual translation systems and compare it to training done on real ground-truth translations. We consider three different dataset sizes for the multilingual setup with $0.5$M, $5$M, and $100$M sentences for each language.

In Table \ref{tab:en-mult}, we see that across different dataset sizes (1.5M = 0.5M for each pair; similar for 15M and 300M) and model complexities, training on synthetic targets outperforms training on ground-truth data. Thus, in the presence of large pre-trained bilingual experts, this provides a recipe for training stronger and more powerful multilingual models of various sizes.

\textbf{German/French/Spanish to English}. We perform similar analysis in the reverse direction where we train multilingual models to translate sentences from German, French and Spanish to English. We use shared tokenizers and vocabularies, with the vocabulary size of 64000, and use bilingual experts trained on large datasets to provide synthetic targets. We train the multilingual models on synthetic and ground-truth targets with $0.5$M, $5$M, and $100$M sentences for each language pair.

We highlight the results in Table \ref{tab:mult-en} for different dataset and model sizes and see consistent improvements on using synthetic targets as opposed to real ground-truth targets for training.

Our experiments on multilingual translation in both directions reveal that when we use bilingual experts for each language pair as teachers, we see consistent improvements when training solely on synthetic data. We believe that this can pave the way for utilization of bilingual models for better and more efficient training of multilingual models, given bilingual models currently outperform multilingual models while the latter are more memory efficient as only a single model needs to be stored.

\vspace{-1mm}
\subsection{Speech Translation}
In end-to-end speech translation (ST), the task is to train a model which translates speech in one language into a text in another. In contrast to text-to-text NMT, data for this particular task is much more scarce and expensive. However, this problem can be solved with the help of readily available text-to-text models and a large corpora of ASR data. Surprisingly, completely replacing all target language transcripts with synthetic data improves the performance of the ST models we trained.

Our speech translation models consist of a $17$-layer Conformer encoder \cite{gulati2020conformer} initialized with pre-trained speech recognition (ASR) encoder followed by a $6$-layer Transformer decoder initialized randomly. For training, we used all available En$\rightarrow$De ST datasets from IWSLT'22 competition~\cite{anastasopoulos2022findings} which amounted to $590$K examples after cleaning. To generate synthetic targets, we used $24 \times 6$ teacher model trained on WMT21 with the optional in-domain fine-tuning on $250$K sentences from Must-C v2 dataset~\cite{cattoni2021must}.

In Table \ref{tab:st}, we see that replacing real targets with synthetic ones leads to over $1.5$ BLEU score improvement, even though the teacher model was trained on out-of-domain data. When teacher model is fine-tuned in-domain, the score improvement goes to $3$ BLEU. Finally, using the synthetic translations, we can expand our dataset by translating ASR-only datasets from IWSLT'22 which do not have German translations. Adding additional $660$K examples leads to another $1.5$ BLEU improvement over, even though the additional data was out of domain to TED talks we evaluate on.

\begin{table}[t]
    \centering
    \small
    \def\arraystretch{1.25}
    \begin{tabular}{c|cccc}
    \toprule
    \textbf{Model} & \textbf{Data} & \textbf{Real} & \textbf{Synthetic} & $\mathbf{\Delta}$ \\
    \midrule
    \multirow{3}{*}{$2\times 2$} & IWSLT & $27.9$ & $30.4$ & $2.5$ \\
    & Medical & $27.3$ & $30.6$ & $3.3$ \\
    & Law & $36.5$ & $39.0$ & $2.5$ \\
    \midrule
    \multirow{3}{*}{$6\times 6$} & IWSLT & $30.0$ & $31.6$ & $1.6$ \\
    & Medical & $30.1$ & $31.7$ & $1.6$ \\
    & Law & $39.8$ & $40.8$ & $1.0$ \\
    \midrule
    \multirow{3}{*}{$24\times 6$} & IWSLT & $31.2$ & $33.1$ & $1.9$ \\
    & Medical & $31.3$ & $33.1$ & $1.8$ \\
    & Law & $40.4$ & $42.1$ & $1.7$ \\
    \bottomrule
    \end{tabular}
    \caption{English to German NMT model trained on WMT21 Machine Translation dataset with real and synthetic targets respectively, evaluated out-of-distribution on different domains. The teacher model used to generate synthetic targets was trained on additionally large amounts of data.}
    \label{tab:en-de-domain}
    \vspace{-5mm}
\end{table}

\vspace{-1mm}
\subsection{Out-of-Domain Evaluation}
One might argue that the model trained on a particular dataset overfits to it and using it to translate sentences from other domains will produce poor results, which can be exacerbated on using synthetic targets from models trained on a particular domain. In the next series of experiments, we evaluate out-of-domain performance of models trained with synthetic data.

Table \ref{tab:en-de-domain} shows the performance of models trained on WMT'21 dataset evaluated on three different non-news domains: TED talks (IWSLT), medical, and law. As we see, training on synthetic targets is beneficial here as well even though all training was done on the WMT dataset.

\vspace{-1mm}
\subsection{In-Domain Fine-tuning}
Finally, we utilize synthetic targets for in-domain fine-tuning, where pre-training is done on real or synthetic data and then further fine-tuning is done using real, out-of-domain synthetic and in-domain synthetic data. In Table \ref{tab:iwslt-ft}, we train NMT models on WMT'21 dataset with either real or synthetic targets and then additionally fine-tune them on IWSLT data which is in-domain to evaluation \texttt{tst-COMMON} dataset comprising of TED talks.

We see that fine-tuning on synthetic targets generated with out-of-domain model actually hurts the model performance but pre-training on it works well. Also, fine-tuning on synthetic targets generated with in-domain model is superior to fine-tuning on real in-domain data no matter what data was used for pre-training. Training from scratch with synthetic targets generated by both models outperforms real targets by a large margin.

\begin{table}[t]
    \centering
    \small
    \def\arraystretch{1.25}
    \begin{tabular}{lccc}
    \toprule
    \textbf{Fine-tuning targets} & $\mathbf{2\times 2}$ & $\mathbf{6\times 6}$ & $\mathbf{24\times 6}$\\
    \midrule
    \multicolumn{4}{c}{\textit{Training from scratch}} \\
    Real & $16.9$ & $17.0$ & $19.2$ \\
    Synthetic, WMT Teacher & $20.6$ & $21.9$ & $22.7$ \\
    \quad + fine-tuned on IWSLT & $21.8$ & $22.9$ & $25.9$ \\
    \midrule
    \multicolumn{4}{c}{\textit{Pre-training on real data}} \\
    Real & $30.9$ & $33.6$ & $34.5$ \\
    Synthetic, WMT Teacher & $30.9$ & $32.6$ & $33.7$ \\
    \quad + fine-tuned on IWSLT & $32.5$ & $34.1$ & $35.5$ \\
    \midrule
    \multicolumn{4}{c}{\textit{Pre-training on synthetic data}} \\
    Real & $32.4$ & $33.8$ & $34.9$ \\
    Synthetic, WMT Teacher & $32.0$ & $33.3$ & $33.6$ \\
    \quad + fine-tuned on IWSLT & $33.4$ & $35.1$ & $35.6$ \\
    \bottomrule
    \end{tabular}
    \caption{English to German NMT model trained on WMT21 dataset and additionally fine-tuned on IWSLT dataset. SacreBLEU on Must-C \texttt{tst-COMMON} dataset is reported.}
    \label{tab:iwslt-ft}
    \vspace{-5mm}
\end{table}


\vspace{-1mm}
\section{Ablations}
\vspace{-1mm}
Our key ablations into understanding the phenomena presented involve analysing whether the results can be solely explained from the lens of optimization, or are there other reasons at play (eg. stochasticity of the predictive model).
\vspace{-1mm}
\subsection{Optimization Problem}
\begin{figure}
    \centering
    \includegraphics[width=\columnwidth]{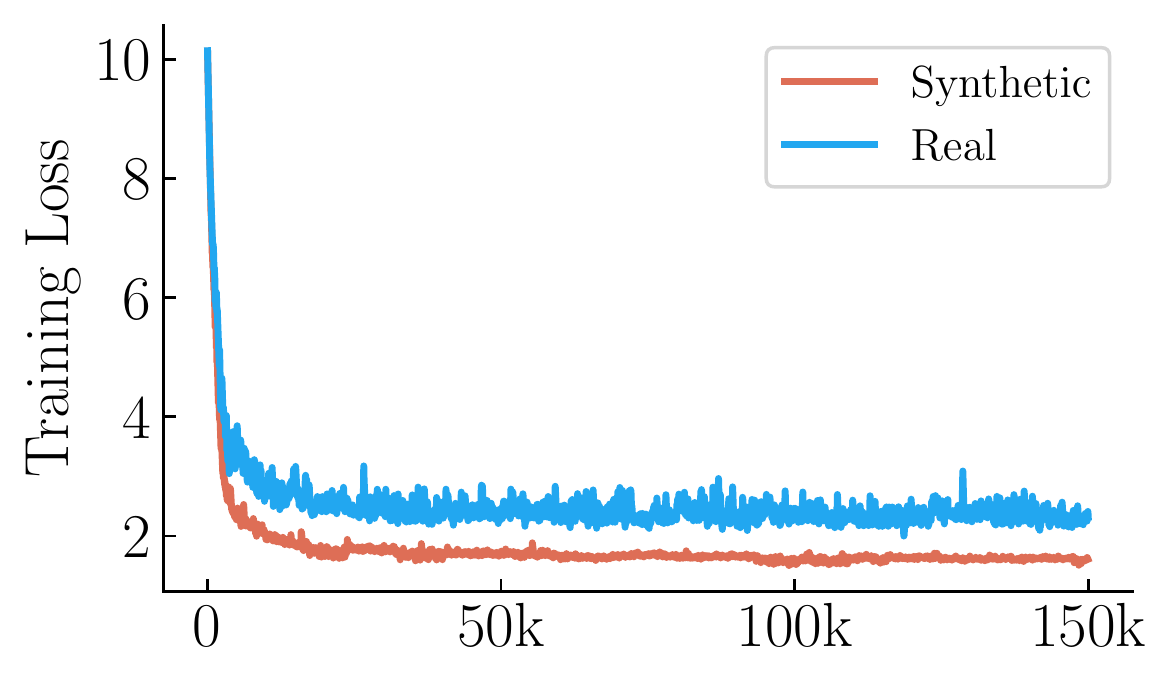}
    \caption{Training loss of models trained with synthetic and real targets on the English to German machine translation. Both models are $6 \times 6$ Transformers; synthetic targets are obtained with a $24 \times 6$ teacher model.}
    \label{fig:train}
    \vspace{-5mm}
\end{figure}

It can be argued that training on teacher outputs is easier \cite{he2019revisiting} and its superior performance is just an artifact of a well-behaved optimization landscape leading to ``better solution spaces''. Figure~\ref{fig:train} shows that synthetic targets are indeed easier to fit as both the training loss and its variance are lower. However, we do not agree that training on synthetic targets leads to better local optima on which real-data training can capitalize on.

To see this, we pre-train some of the systems from Section~\ref{sec:exp} on synthetic targets and half-way during the training, switch out to real ground-truth targets. Our hypothesis is that if training on synthetic targets leads to better solution spaces, using the corresponding model parameters as initialization will not hurt the overall performance of training with real data, as evaluation data is considered to be from the same distribution as the real data and not the synthetic one.

\begin{figure*}
\centering
\includegraphics[width=\textwidth]{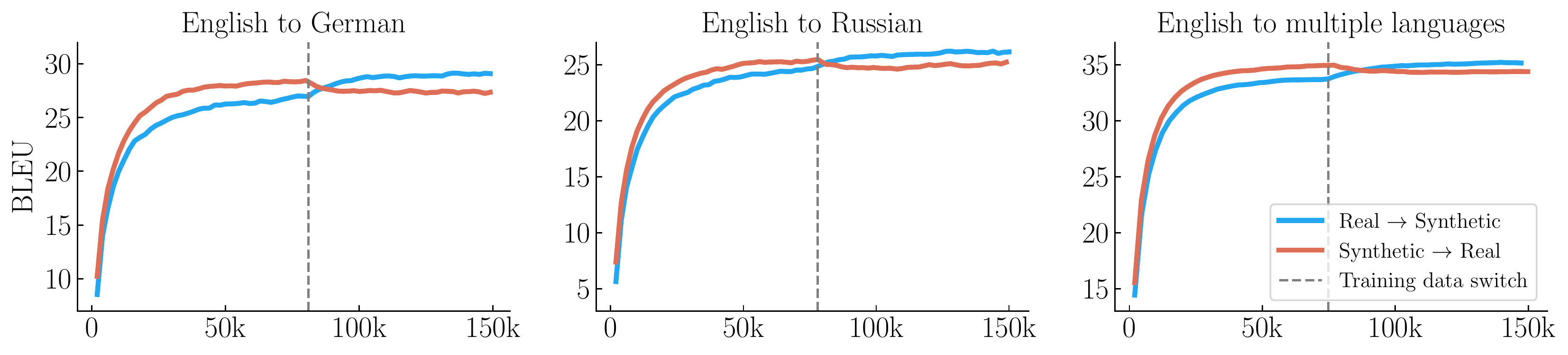}
\caption{We highlight that the benefits of training using a teacher model are not because of optimization. In particular, we train models on synthetic targets and then mid-way between training, we switch to real targets. We see that this leads to a degradation in performance implying that synthetic training does not lead to better initialization for the model, and this is consistent over different setups. Further, if we train on real targets and switch to synthetic ones, we see an immediate boost in performance.}
\label{fig:opt}
    \vspace{-5mm}
\end{figure*}

However, our experiments in Figure \ref{fig:opt} imply that when we switch the data from synthetic to real, we do see an immediate drop in performance. Moreover, we also see that when we train on real targets and switch to synthetic targets, we get an immediate boost in performance. Together, the two results indicate that training with synthetic data does not leads to better solutions to the underlying optimization problem and that the reasoning behind its workings is something else.

\vspace{-1mm}
\subsection{Top-k Performance}
Next, we analyze whether the models trained with synthetic targets are more deterministic than those trained with real data. Our hypothesis is that training on synthetic targets dampens some of the noise present in the ground-truth dataset. To explore this claim, we use the models trained in Section \ref{sec:exp} and evaluate them at different levels of top-k sampling for inference and compare it with models trained on ground-truth data.

\begin{figure}
    \centering
    \includegraphics[width=\columnwidth]{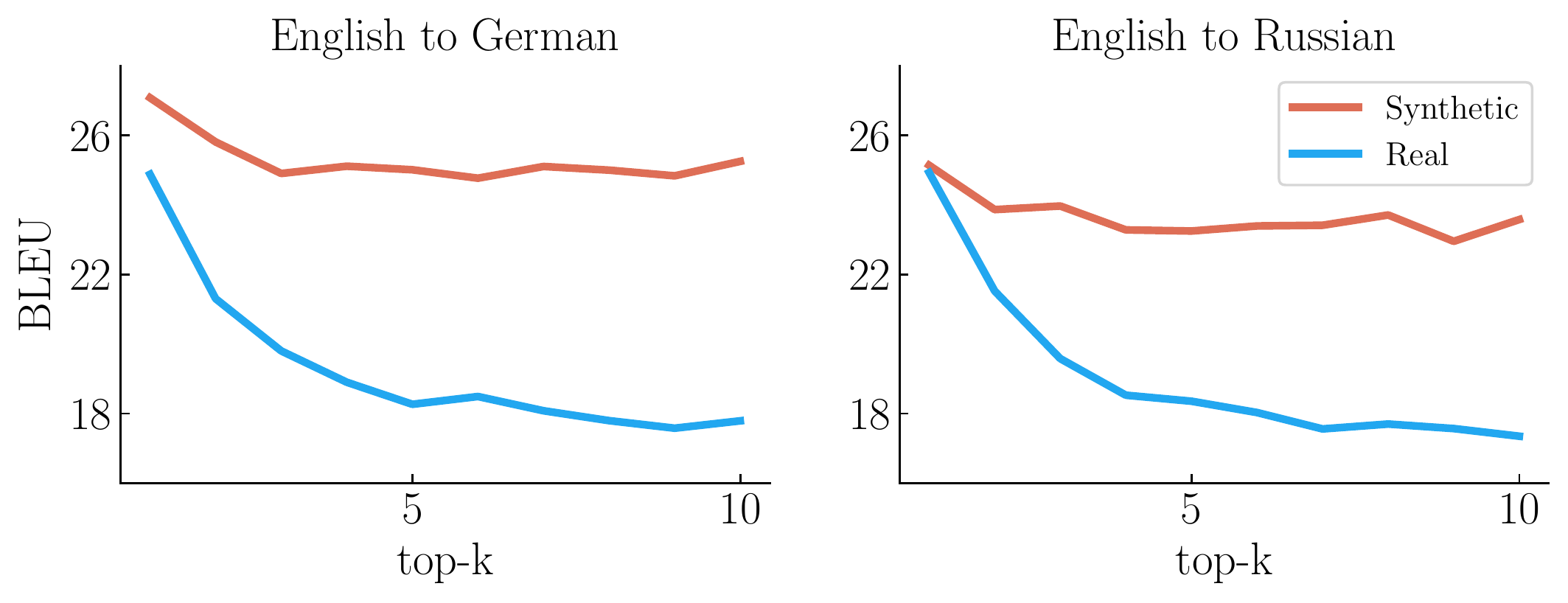}
    \caption{Degradation of Machine Translation performance with increasing k in top-k sampling for the decoding procedure for both English to German and English to Russian translations using student $24\times 6$ models trained with real and synthetic targets respectively.}
    \label{fig:topk}
    \vspace{-6mm}
\end{figure}

Our findings in Figure \ref{fig:topk} showcase that the drop in performance with increasing $k$ in top-k is less in models trained with synthetic targets than in those trained with real targets. This does highlight that models trained with synthetic targets capture less noise as the degradation of performance is less when we make the sampling more noisy.

\begin{table}[t]
    \centering
    \small
    \def\arraystretch{1.25}
    \begin{tabular}{ccccc}
    \toprule
    \multirow{2}{*}{\textbf{Language}} & \multirow{2}{*}{\textbf{Targets}} & \multicolumn{3}{c}{\textbf{Predictive Entropy}} \\
    & & $\mathbf{2\times 2}$ & $\mathbf{6\times 6}$ & $\mathbf{24\times 6}$ \\
    \midrule
    \multirowcell{2}{English to \\ German} & Real & $2.4$ & $2.4$ & $2.5$ \\
     & Synthetic & $1.8$ & $1.8$ & $1.7$ \\
    \midrule
    \multirowcell{2}{English to \\ Russian} & Real & $2.6$ & $2.4$ & $2.3$ \\
     & Synthetic & $1.8$ & $1.8$ & $1.9$ \\
     \bottomrule
    \end{tabular}
    \caption{Average entropy of the predictive distribution for translation $p(\vy|\vx)$ is lower for models trained with synthetic targets than those trained with real targets, implying more deterministic nature of the former.}
    \label{tab:entropy}
    \vspace{-5mm}
\end{table}
\vspace{-1mm}
\subsection{Predictive Entropy}
As a final analysis of whether models trained with synthetic targets are more deterministic, we compute the predictive entropy of the distribution over the logits that predict the next token in translation. Our findings in Table \ref{tab:entropy} highlight that indeed the predictive entropy of models trained with synthetic targets is lower than the models trained with ground-truth targets, implying more deterministic nature of the former translation systems.

Together, our analysis on top-k performance and predictive entropy provide some evidence that the models trained with synthetic targets are more deterministic and hence are more robust and perform better, even on out of distribution shifts.
\vspace{-1mm}
\section{Conclusion}
\vspace{-1mm}
Inspired by the recent advances in knowledge distillation and the need for better performing low-resource and low-compute models, we provide a recipe that leverages large-scale pre-trained translation systems as teacher models which provide synthetic targets for training of smaller and low-resource models. Surprisingly, we see a considerable increase in the performance of smaller models when only teacher outputs are provided as opposed to any proportion of real ground-truth translations. We also see additional benefits of using synthetic targets for training, namely faster convergence and improved translations with top-k sampling when compared to models trained solely on real ground-truth translations.

This improvement in performance of small or low-resource models comes at the additional inference costs tied with the large teacher model. However, this is a one-time data curating cost based on which training of multiple different smaller models can be accomplished which will enjoy not only faster inference (owing to their smaller size) but also better performance than if they were trained on the original data, sometimes even better than larger models trained on real data. We believe this is an exciting avenue of research as low-compute high-performance models are important when deployed in constrained settings like edge computing and mobile devices.
\section*{Ethics Statement}
We provide a methodology for improving the performance of resource and data constrained translation systems which rely on obtaining synthetic targets from larger pre-trained systems. Given the dependence on large pre-trained systems, we believe that their biases can negatively impact the biases and fairness of the smaller consecutively trained systems. This is a problem common with any type of knowledge transfer where the biases of the base model can also be transferred to the student system and approaches on mitigating biases in the larger models in the first place would be a potential solution that can alleviate this problem.
\clearpage
\bibliography{custom}
\bibliographystyle{acl_natbib}
\clearpage
\appendix
\section*{\LARGE Appendix}
\section{Implementation Details}

For all our experiments, we rely on the NeMo codebase published in \citet{kuchaiev2019nemo}. We do not perform extensive hyperparameter selection and instead just rely on the defaults provided. All the models that we train from scratch use the pre layer-norm transformer variant \cite{xiong2020layer} and are trained with $0.001$ learning rate with a linear warmup followed by an exponential decay. For all the synthetic targets, we use beam search with beam size $4$ for generating translations. All experiments also use label smoothing of $0.1$ with a dropout of $0.1$ as well. We only vary the models in their depth, while keeping the attention layer of $512$ dimensions, feed-forward residual connections of $2048$ and $8$ attention heads, as is typical with transformer models. All experiments were done on $16$ GPUs and for $150,000$ iterations by when convergence of all models had been achieved. All the results are reported by considering single runs of the model as our experiments revealed very low variance between different runs.

\end{document}